\documentclass[10pt]{article}

\usepackage[letterpaper, margin=2.54cm, top=2.54cm]{geometry}
\usepackage[super,comma,sort&compress]{natbib}
\usepackage{lmodern}
\usepackage{authblk} 
\usepackage{amssymb,amsmath}
\usepackage{wrapfig}
\usepackage{graphicx,grffile}
\usepackage[labelfont=bf,labelsep=period]{caption}
\usepackage{ifxetex,ifluatex}
\usepackage{placeins}
\ifnum 0\ifxetex 1\fi\ifluatex 1\fi=0 
  \usepackage[T1]{fontenc}
  \usepackage[utf8]{inputenc}
\else 
  \ifxetex
    \usepackage{mathspec}
  \else
    \usepackage{fontspec}
  \fi
  \defaultfontfeatures{Ligatures=TeX,Scale=MatchLowercase}
\fi
\IfFileExists{upquote.sty}{\usepackage{upquote}}{}
\IfFileExists{microtype.sty}{%
	\usepackage{microtype}
	\UseMicrotypeSet[protrusion]{basicmath} 
}{}

\usepackage{lipsum} 

\makeatletter
\renewcommand\subsubsection{\@startsection{subsubsection}{3}{\z@}%
	{-3.25ex\@plus -1ex \@minus -.2ex}%
    {-1.5ex \@plus -.2ex}
    {\normalfont\itshape}}
\makeatother

\makeatletter 
\renewcommand\@biblabel[1]{#1.} 
\makeatother %

\usepackage{etoolbox}
\makeatletter
\patchcmd{\@maketitle}{\LARGE}{\bfseries\fontsize{15}{16}\selectfont}{}{}
\makeatother

\makeatletter
\def\maxwidth{\ifdim\Gin@nat@width>\linewidth\linewidth\else\Gin@nat@width\fi}
\def\maxheight{\ifdim\Gin@nat@height>\textheight\textheight\else\Gin@nat@height\fi}
\makeatother

\setkeys{Gin}{width=\maxwidth,height=\maxheight,keepaspectratio}
\ifx\paragraph\undefined\else
\let\oldparagraph\paragraph
\renewcommand{\paragraph}[1]{\oldparagraph{#1}\mbox{}}
\fi
\ifx\subparagraph\undefined\else
\let\oldsubparagraph\subparagraph
\renewcommand{\subparagraph}[1]{\oldsubparagraph{#1}\mbox{}}
\fi
\usepackage{hyperref}
\hypersetup{
	unicode=true,
	pdftitle={Improving generalization in drug response prediction via independent autoencoder pretraining},
	pdfauthor={Author One, Author Two, Author Three},
	pdfkeywords={keyword1, keyword2},
	pdfborder={0 0 0},
	breaklinks=true
}
\title{\vspace{-2em} Sample-Efficient Adaptation of Drug-Response Models to Patient Tumors under Strong Biological Domain Shift}

\author{\bf\fontsize{13}{14}\selectfont Camille Jimenez Cortes, Philippe Lalanda, German Vega\\
Université Grenoble Alpes (UGA), France}

\date{} 

\begin{document}
\maketitle
\vspace{-4em} 

\section*{Abstract}
\label{abstract}

Predicting drug response in patients from preclinical data remains a major challenge in precision oncology due to the substantial biological gap between in vitro cell lines and patient tumors. Rather than aiming to improve absolute in vitro prediction accuracy, this work examines whether explicitly separating representation learning from task supervision enables more sample-efficient adaptation of drug-response models to patient data under strong biological domain shift. We propose a staged transfer-learning framework in which cellular and drug representations are first learned independently from large collections of unlabeled pharmacogenomic data using autoencoder-based representation learning. These representations are then aligned with drug-response labels on cell-line data and subsequently adapted to patient tumors using few-shot supervision. Through a systematic evaluation spanning in-domain, cross-dataset, and patient-level settings, we show that unsupervised pretraining provides limited benefit when source and target domains overlap substantially, but yields clear gains when adapting to patient tumors with very limited labeled data. In particular, the proposed framework achieves faster performance improvements during few-shot patient-level adaptation while maintaining comparable accuracy to single-phase baselines on standard cell-line benchmarks. Overall, these results demonstrate that learning structured and transferable representations from unlabeled molecular profiles can substantially reduce the amount of clinical supervision required for effective drug-response prediction, offering a practical pathway toward data-efficient preclinical-to-clinical translation.

\section{Introduction}

Predicting drug response from molecular profiles is a central goal of precision oncology, with the promise of guiding therapeutic decisions based on tumor-specific characteristics. Large-scale pharmacogenomic screening efforts on cancer cell lines have enabled the development of increasingly accurate machine-learning models by pairing molecular features with measured drug sensitivities \cite{cai2022machine}. These resources have become a cornerstone for computational drug-response prediction, providing standardized, data-rich environments for model training and benchmarking.

Despite this progress, translating models trained on in vitro cell lines to patient tumors remains a major challenge. Cell lines represent simplified biological systems that differ substantially from primary tumors in terms of cellular heterogeneity, microenvironmental context, and clinical confounders \cite{elnaqa2023assessment}. As a result, predictive models that perform well on cell-line benchmarks often exhibit limited generalization when applied directly to patient data. This preclinical-to-clinical gap continues to constrain the practical impact of drug-response prediction (DRP) models in real-world oncology settings.

From a machine-learning perspective, this limitation can be framed as a problem of domain shift. Models are typically trained in a source domain defined by cell-line molecular profiles and screening conditions, while deployment occurs in a target domain corresponding to patient tumors. Differences in data distributions between these domains—arising from biological variability, experimental protocols, and measurement platforms—violate the assumptions under which standard supervised learning guarantees generalization \cite{blanchard2011generalizing,blanchard2021marginal}. Addressing such shifts is therefore essential for reliable patient-level prediction.

A growing body of work has explored transfer learning and domain adaptation (DA) strategies for DRP, including adversarial alignment, discrepancy minimization, and representation learning approaches \cite{ganin2015unsupervised,jiang2023transcdr}. Among these, autoencoder-based methods have attracted particular interest due to their ability to learn compact latent representations from high-dimensional molecular data \cite{partin2023review}. Several models jointly encode cellular omics and drug features using autoencoders coupled with supervised predictors, achieving strong performance on cell-line benchmarks. For example, DeepDRA integrates multi-omics cell profiles with drug descriptors and molecular fingerprints through modality-specific autoencoders trained end-to-end with a prediction network \cite{mohammadzadeh2024deepdra}. However, such single-phase training strategies tightly couple representation learning and task supervision and rely almost exclusively on labeled cell--drug pairs, thereby limiting the exploitation of large collections of unlabeled molecular data.

In clinical settings, labeled patient data are inherently scarce, making it unrealistic to expect large supervised training sets in the target domain. Consequently, the key practical question is not whether a model achieves state-of-the-art performance when trained and evaluated on cell lines, but whether it can be adapted efficiently to a new biological domain using only a small number of labeled patient samples. Few-shot learning provides a natural framework to address this challenge by enabling rapid adaptation of pretrained models with minimal supervision \cite{ma2021fewshot}. While few-shot approaches have shown promise for DRP in patient-derived tumor cells and xenograft models, it remains unclear under which conditions representation learning genuinely improves adaptation efficiency rather than merely enhancing in-domain accuracy.

In this work, we investigate whether explicitly separating representation learning from task supervision can enable sample-efficient adaptation of DRP models to new domains, even when direct in vitro performance remains comparable to existing approaches. Our focus is therefore not on maximizing in-domain accuracy on cell-line benchmarks, but on minimizing the amount of target-domain supervision required to achieve reliable adaptation under strong biological domain shift. We hypothesize that unsupervised pretraining on large collections of unlabeled molecular profiles can yield structured and transferable representations that accelerate learning in the target domain, while offering limited benefit when source and target distributions overlap substantially \cite{theodoris2023transfer}.

To test this hypothesis, we propose a staged transfer-learning framework based on autoencoder-driven representation learning and evaluate it across a spectrum of settings characterized by increasing levels of distribution shift, ranging from in-domain leave-out protocols to cross-dataset transfer and patient-level few-shot adaptation. Through this analysis, we aim to clarify when representation learning improves DRP in practice and to quantify its impact on data efficiency under clinically realistic constraints.

\paragraph{Contributions and paper organization.}
This work makes three main contributions. First, we propose a staged transfer-learning framework that explicitly separates unsupervised representation learning, task-specific alignment, and few-shot clinical adaptation for DRP. Second, we demonstrate through a systematic experimental analysis that, although unsupervised pretraining yields limited gains for direct in vitro prediction, it substantially improves few-shot adaptation to patient tumors, reducing the number of labeled target samples required for effective transfer. Third, we link these performance patterns to latent-space geometry, providing mechanistic insight into when and why representation learning is beneficial under strong biological domain shift.

The remainder of the paper is organized as follows. Section~2 reviews related work on DRP and domain shift. Section~3 presents the proposed framework. Section~4 describes the experimental setup. Section~5 reports the results, and Section~6 discusses their implications, limitations, and future directions.

\section{Related Work}
A central challenge in applying machine-learning models to biomedical data is their sensitivity to changes in data distribution. Models are typically trained in controlled, data-rich environments (source domains) but are often deployed in settings where data distributions differ substantially (target domains), such as independent cohorts or patient populations. These discrepancies, commonly referred to as domain shift, arise from biological heterogeneity, experimental protocols, and population-specific effects, and can severely degrade predictive performance.

Formally, a domain is defined as a joint distribution $D = P(X,Y)$ over inputs $X$ and labels $Y$ \cite{blanchard2011generalizing,blanchard2021marginal}. Domain shift occurs when source and target domains differ, i.e., $P(X_S,Y_S) \neq P(X_T,Y_T)$, due to changes in input distributions (covariate shift), label distributions (label shift), or their conditional relationship (concept shift) \cite{ben2010theory,zhao2019invariant}. In biomedical applications, such shifts are particularly pronounced and often unavoidable.

DRP exemplifies this problem. Models trained on large-scale in vitro pharmacogenomic screens operate in a domain that differs fundamentally from that of patient tumors. While cell-line datasets provide abundant and standardized molecular profiles, patient data introduce additional sources of variability, including tumor microenvironment effects, intra-tumor heterogeneity, and clinical confounders. As a result, models optimized on cell-line data frequently fail to generalize to patient cohorts, motivating approaches explicitly designed to cope with domain shift.

Two methodological frameworks have been widely studied to address this issue: DA and Domain Generalization (DG). DA focuses on transferring knowledge from source domains to a specific target domain that is accessible during training, often under limited or absent target labels \cite{ben2006analysis,ganin2015unsupervised}. In contrast, DG aims to learn models that generalize to unseen target domains by exploiting variability across multiple source domains \cite{blanchard2011generalizing,muandet2013domain}. Both frameworks seek representations that capture task-relevant information while discarding domain-specific variability, a goal that is particularly challenging in high-dimensional biomedical settings and naturally motivates representation learning approaches.

In this work, we focus on a DA setting, in which the target domain (patient tumors) is known at training time but labeled data are extremely limited. As such, our objective is not to learn domain-invariant representations for unseen targets, as in DG, but to enable sample-efficient adaptation under strong biological domain shift. Importantly, this setting does not assume explicit distribution alignment between source and target domains, but rather emphasizes rapid specialization using limited target-domain supervision.

Within DA and DG, existing approaches can be broadly grouped into three families. Statistical alignment methods reduce discrepancies between domains by minimizing divergence measures such as Maximum Mean Discrepancy or covariance differences \cite{tzeng2014deep,long2014transfer,sun2016coral}. Adversarial approaches enforce domain invariance through domain discriminators trained in opposition to feature extractors, as in Domain-Adversarial Neural Networks \cite{ganin2015unsupervised}. Reconstruction-based approaches rely on autoencoder architectures to learn latent representations that preserve essential input structure while attenuating domain-specific noise \cite{bousmalis2016dsn,ghifary2016drcn,zhuang2017tlda}. Among these, autoencoder-based methods are particularly attractive for biomedical data due to their ability to handle high dimensionality and heterogeneous modalities.

Autoencoders have been widely adopted for multi-omics integration and DRP. Early studies showed that autoencoders can effectively compress molecular profiles into informative latent spaces \cite{tan2015autoencoder,chaudhary2018deep}. Subsequent work extended this paradigm by jointly encoding cellular omics and drug features to model cell--drug interactions, including methods such as MOLI, DeepCDR, and DeepDRA \cite{sharifi2019moli,zhang2021deepcdr,mohammadzadeh2024deepdra}. In particular, DeepDRA employs modality-specific autoencoders trained jointly with a prediction network in a single-phase supervised setting, learning representations directly optimized for DRP.

While these approaches demonstrate strong performance on cell-line benchmarks, their tightly coupled training strategy limits the exploitation of unlabeled molecular data and may reduce robustness under strong domain shift. Representation learning and task supervision are learned simultaneously from labeled cell--drug pairs, which constrains the diversity of biological states captured in the latent space and may hinder transfer to domains far from the training distribution, such as patient tumors.

More recent studies have explored explicit DA strategies for DRP, including adversarial alignment and discrepancy minimization between pharmacogenomic datasets or between preclinical and clinical data \cite{jiang2023transcdr,he2022context}. Although these methods can improve cross-domain performance, they often require access to target-domain data during training and rely on assumptions about label distributions that may not hold in clinical settings.

Despite these advances, few works explicitly investigate when representation learning is beneficial under domain shift or disentangle the respective roles of representation learning and task supervision. This gap motivates training paradigms that separate unsupervised representation learning from downstream supervision, leverage large collections of unlabeled molecular profiles, and enable sample-efficient adaptation to patient data under strong biological shift.

\section{Proposal}

As discussed in the previous sections, DRP models trained on in vitro cell-line data are exposed to a strong domain shift when applied to patient tumors. While large pharmacogenomic screens provide abundant molecular profiles and drug-response measurements, patient-level data remain scarce and heterogeneous. This discrepancy limits the direct applicability of fully supervised models and motivates training strategies that explicitly account for distribution shifts while minimizing clinical supervision requirements.

To address this challenge, we propose \textbf{STaR-DR (Staged Transfer of Representations for Drug Response)},a staged transfer learning framework that explicitly separates representation learning, task alignment, and clinical adaptation. The core idea is to exploit large collections of unlabeled molecular data to learn structured and transferable representations of cells and drugs, to align these representations with pharmacological signal using cell-line supervision, and finally to adapt the resulting model to the patient domain using limited labeled clinical data.

The proposed framework is composed of three main components: (i) a cell encoder that maps high-dimensional molecular profiles of cancer cells to a compact latent representation, (ii) a drug encoder that embeds molecular descriptors and fingerprints into a separate latent space, and (iii) a lightweight prediction head that combines these representations to estimate drug sensitivity. Rather than training these components jointly in a single supervised phase, the model follows a three-stage training strategy designed to improve robustness under strong domain shift.

In the first phase (P1), the cell and drug encoders are pretrained independently using autoencoders in a self-supervised manner. This phase relies exclusively on unlabeled molecular data and aims to learn latent spaces that capture fundamental biological and chemical variability while remaining agnostic to the downstream prediction task. By using reconstruction as the learning signal, the encoders are encouraged to model intrinsic properties of cells and drugs instead of dataset-specific correlations that may not transfer across domains.

\begin{figure}[t]
    \centering
    \includegraphics[width=0.85\linewidth]{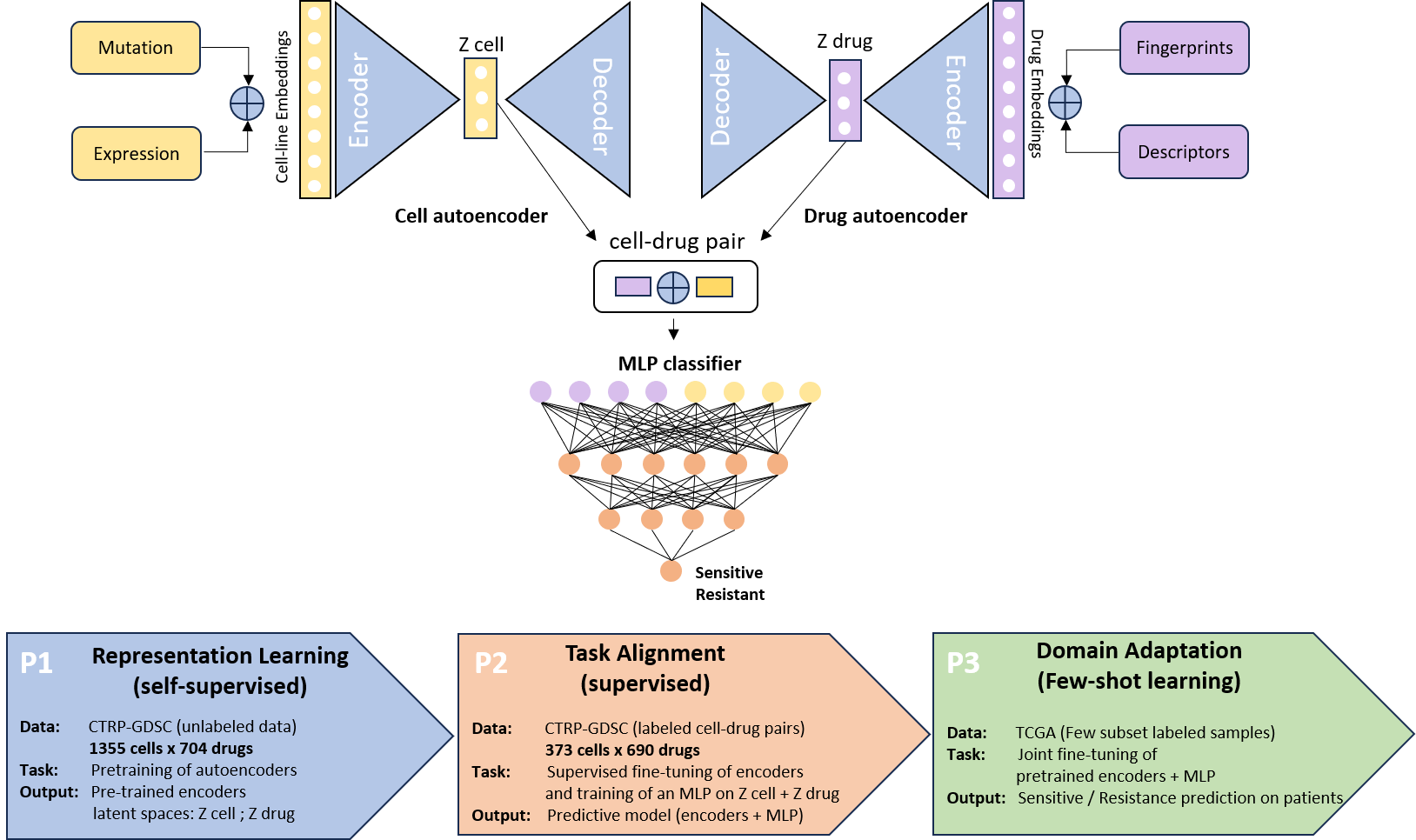}
    \caption{\textbf{STaR-DR framework.}
Cell and drug features are independently encoded into latent representations and combined for drug-response prediction (DRP). Training includes unsupervised pretraining on CTRP–GDSC, supervised alignment on cell-line response data, and few-shot adaptation to TCGA.}
    \label{fig:overview}
\end{figure}

In the second phase (P2), the pretrained encoders are jointly fine-tuned with a lightweight classifier using labeled cell--drug response pairs from large-scale pharmacogenomic screens. This step aligns the latent representations with pharmacological signal while preserving the structure learned during unsupervised pretraining. Importantly, the prediction head is intentionally kept simple, so that performance gains primarily reflect the quality and transferability of the learned representations rather than increased model complexity.

In the third phase (P3), the model is adapted to the clinical setting using a few-shot learning strategy on patient data. A small number of labeled patient--drug response pairs are used to specialize the pretrained and task-aligned model to the patient domain. Given the strong biological shift between cell lines and tumors, adaptation primarily focuses on the cellular representation, while drug representations are largely preserved to avoid overfitting in the presence of limited clinical data. This design enables sample-efficient patient-level adaptation without requiring explicit domain-alignment constraints or extensive retraining.

An overview of the proposed framework and its three training stages is illustrated in Figure~\ref{fig:overview}. Overall, this staged design enables STaR-DR to leverage unlabeled molecular data that would otherwise remain unused, mitigate the impact of strong distribution shifts between preclinical and clinical settings, and support robust and sample-efficient adaptation to patient data.


\section{Experimental setup}

This section describes the datasets, modeling choices, and evaluation protocols used to assess the proposed staged transfer-learning framework under increasing levels of distribution shift.

\subsection{Datasets and modalities}

We consider three datasets reflecting progressively stronger domain shifts. Two large-scale preclinical cell-line resources are used for training and validation: CTRP–GDSC \cite{rees2016ctrp,yang2013gdsc}, which serves as the primary training dataset, and CCLE \cite{barretina2012ccle}, used as an independent benchmark for cross-dataset evaluation. Patient-level transfer is assessed on TCGA \cite{liu2018tcga}.
CTRP and GDSC pair molecular profiles of cancer cell lines with drug sensitivity measurements, enabling supervised learning on cell–drug response pairs. CCLE provides complementary multi-omics characterizations of cancer cell lines collected under distinct experimental protocols, making it suitable for evaluating cross-dataset generalization within the in vitro domain. In contrast, TCGA comprises multi-omics and clinical data from patient tumors and is used to assess adaptation under strong preclinical-to-clinical domain shift.

Processed molecular features and binarized drug-response labels were obtained from the public DeepDRA repository. Cellular profiles comprise gene expression and somatic mutation data, while drug representations include molecular descriptors and structural fingerprints. Gene expression features are represented as continuous values per gene, capturing the transcriptional state of each sample, whereas somatic mutations are encoded as binary indicators at the gene level (1 for mutated, 0 for wild-type).
On the drug side, molecular structure is represented using Morgan fingerprints, encoded as high-dimensional binary vectors capturing circular substructures, together with continuous physicochemical and topological molecular descriptors. Modality-specific feature matrices are aligned by intersecting common entities across datasets and concatenated to form the final cell and drug input spaces.

After preprocessing and alignment, CTRP–GDSC contains 373 cell lines and 690 drugs (28{,}833 cell--drug pairs), CCLE contains 470 cell lines and 23 drugs (633 pairs), and TCGA includes 714 patients and 32 drugs (2{,}485 pairs), with gene expression as the only available molecular modality. To ensure cross-domain comparability, the feature schema learned on CTRP–GDSC is persisted and reused to reindex CCLE and TCGA. Drug-response labels are binarized, with resistant pairs encoded as class~0 and sensitive pairs as class~1.

\subsection{Preprocessing}

All features are independently normalized using per-feature min--max scaling, and missing values are imputed with zeros. During training, an additional normalization step is applied using statistics computed on the training split only and reused for validation and testing to prevent data leakage.

Class imbalance is addressed exclusively at training time using random undersampling. In the CTRP–GDSC dataset, drug–response labels are moderately imbalanced, with 10,057 sensitive and 18,776 resistant cell–drug pairs. After a stratified train–validation split, undersampling is applied to the training subset only to balance class proportions, while validation and test sets retain the original label distribution to ensure unbiased performance evaluation.

\subsection{Model architecture}

The model consists of two modality-specific autoencoders—one for cellular features and one for drug features—followed by a lightweight multilayer perceptron (MLP) classifier operating on the concatenated latent representations. Latent dimensionalities are set to 700 for cells and 50 for drugs, reflecting the higher complexity of multi-omics cellular profiles relative to drug features. Autoencoders are trained using mean squared error reconstruction loss and optimized with Adam (learning rate $10^{-3}$, weight decay $10^{-8}$). The prediction head is a one-hidden-layer MLP with 128 ReLU units, followed by a sigmoid output layer optimized using binary cross-entropy (BCE) loss. All models are implemented in PyTorch, trained with a batch size of 64 for 25 epochs, and random seeds are fixed to 42 to ensure reproducibility. 

\subsection{Training protocol}

Training follows the staged framework described earlier. In Phase~1, cell and drug autoencoders are pretrained independently on CTRP–GDSC using unlabeled data to learn transferable representations. In Phase~2, pretrained encoders are jointly fine-tuned with the MLP classifier on labeled CTRP–GDSC cell--drug pairs to align representations with pharmacological signal. In Phase~3, the resulting model is adapted to TCGA using a few-shot learning protocol. Adaptation primarily targets the cellular encoder, while the drug encoder is kept fixed to avoid overfitting given the limited number of compounds in TCGA.

\subsection{Baseline model}

For comparison, we implement a single-phase supervised autoencoder--MLP baseline adapted from DeepDRA. This baseline matches the proposed framework in architecture and training settings but is trained end-to-end on labeled CTRP–GDSC data without unsupervised pretraining or patient-level adaptation. To ensure a fair comparison, data normalization and class balancing are performed strictly within the training split. We refer to this baseline as AE--MLP.

We deliberately restrict comparisons to a single-phase autoencoder--MLP baseline (AE--MLP) that matches the proposed framework in architecture, optimization, and training protocol. This design choice isolates the effect of the training strategy---namely the explicit separation between representation learning and task supervision---from confounding factors related to model capacity or architectural differences. Comparisons with heterogeneous state-of-the-art models would primarily reflect differences in architecture, modality coverage, or optimization, rather than the specific contribution of staged representation learning to adaptation efficiency, which is the focus of this study.

\subsection{Evaluation protocol and metrics}

Model performance is evaluated using both pair-level and group-aware protocols. In addition to standard pair-level splits, we employ Leave-Cell-Out (LCO) and Leave-Drug-Out (LDO) settings to assess generalization to unseen cell lines and novel compounds, respectively. Performance is reported using ROC--AUC and PR--AUC, which provide threshold-free assessments robust to class imbalance. Balanced accuracy is additionally reported for in-domain and cross-dataset evaluations.

\section{Results}

We evaluate the proposed staged framework across a range of experimental settings designed to progressively increase the magnitude of distribution shift between training and evaluation data. Rather than focusing solely on absolute in-domain performance, our analysis emphasizes adaptation efficiency, i.e., how rapidly performance improves as limited labeled data become available in the target domain.

\subsection{In-domain robustness under leave-out protocols}

We first evaluate in-domain generalization on CTRP--GDSC using three complementary cross-validation protocols designed to probe different generalization regimes: (i) a standard pair-level split, (ii) LCO, in which cell lines are disjoint between training and validation sets, and (iii) LDO, in which compounds are disjoint between training and validation sets.

\begin{figure}[t]
    \centering
    \includegraphics[width=0.95\linewidth]{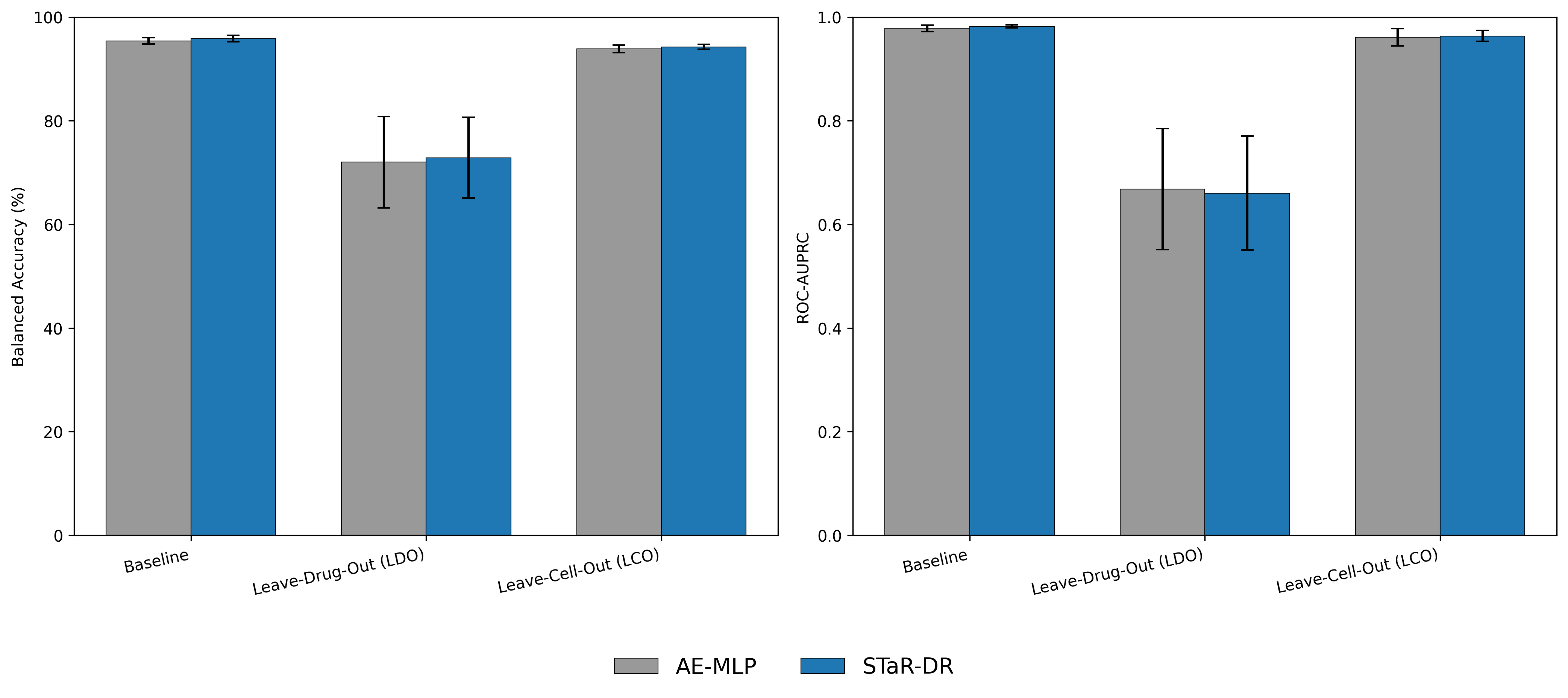}
    \caption{\textbf{In-domain cross-validation performance under leave-out protocols.}
Five-fold cross-validation on the CTRP--GDSC dataset using cell-line gene expression and mutation profiles combined with drug descriptors and Morgan fingerprints. Bars report mean $\pm$ s.d. balanced accuracy (left) and area under the precision--recall curve (AUPRC, right) across three evaluation settings: standard pair-level split (Baseline), Leave-Drug-Out (LDO), and Leave-Cell-Out (LCO).}

    \label{fig:in-domain}
\end{figure}

As shown in Figure~\ref{fig:in-domain}, STaR-DR and AE-MLP achieve similarly high performance under the standard pair-level split, indicating comparable fitting capacity when training and evaluation distributions are closely aligned. This result confirms that separating representation learning from task supervision does not inherently improve in-domain performance when sufficient labeled cell--drug pairs are available.

Under the LCO protocol, performance remains close to the pair-level baseline for both methods, indicating that generalization to unseen cell lines within the same pharmacogenomic dataset is relatively straightforward. This behavior is largely structural: in CTRP–GDSC, drug-response patterns are predominantly driven by drug-specific effects, which are largely conserved across cell lines. As a result, holding out cell lines preserves the dominant drug-related signal, leading to limited effective distribution shift despite the absence of specific cellular profiles during training.

In contrast, performance degrades substantially under LDO for both approaches, highlighting the intrinsic difficulty of extrapolating drug-response predictions to previously unseen compounds. By removing entire drugs from the training set, LDO suppresses the dominant source of predictive structure, forcing models to extrapolate beyond the observed chemical space. Importantly, no systematic performance gap is observed between the staged framework and the single-phase baseline across any of the three in-domain protocols, indicating that unsupervised pretraining does not provide a consistent advantage for direct in vitro prediction or for generalization within closely related cell-line domains.

Together, these results establish a critical reference point for the remainder of the study: the primary contribution of the proposed framework does not lie in improving absolute performance on cell-line benchmarks, but rather in its ability to support more efficient adaptation under stronger domain shift, as examined in subsequent cross-dataset and patient-level evaluations.

\begin{figure}[t]
    \centering
    \includegraphics[width=0.95\linewidth]{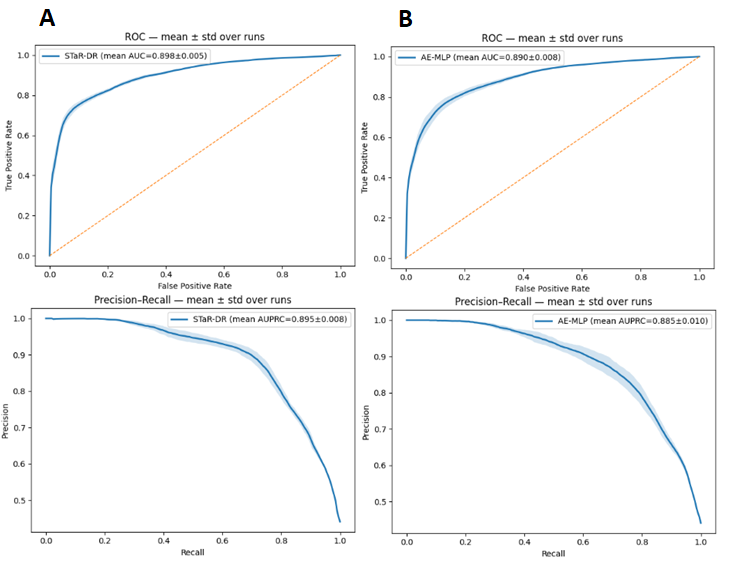}
    \vspace{0.75em}
    \includegraphics[width=0.35\linewidth]{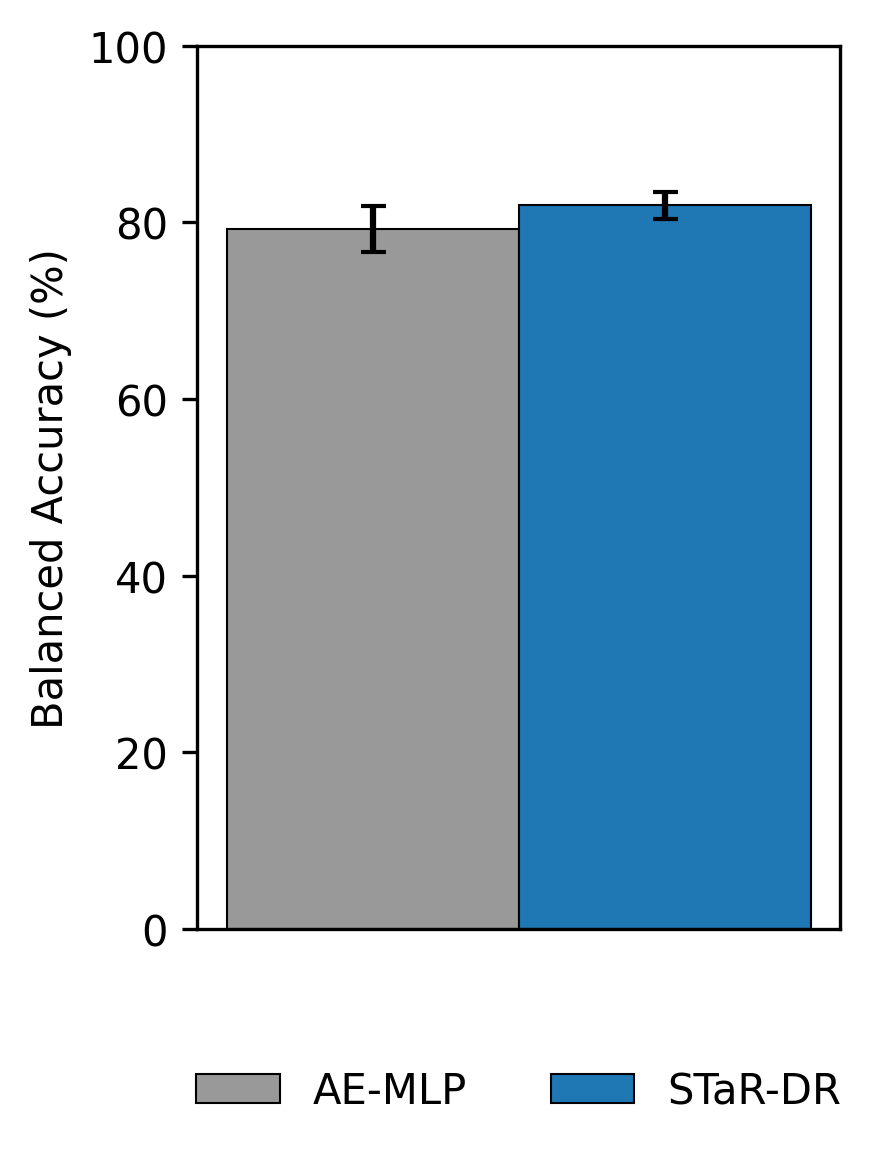}
    \caption{\textbf{Cross-dataset performance on CCLE.}
    Models trained on CTRP--GDSC are evaluated on CCLE. (A--B) ROC (top) and Precision--Recall (bottom) curves averaged over 10 runs. (A) STAR-DR. (B) Single-phase baseline (AE-MLP). Shaded regions indicate $\pm$1 s.d. (C) Balanced accuracy (mean $\pm$ s.d.).}
    \label{fig:cross_dataset}
\end{figure}

\FloatBarrier

\subsection{Cross-dataset generalization to CCLE}

We next assess cross-dataset generalization by training models on CTRP--GDSC and evaluating them on CCLE. This setting introduces a moderate level of distribution shift arising from differences in cell-line panels, experimental protocols, and data preprocessing pipelines, while remaining confined to the in vitro cell-line domain.

As shown in Figure~\ref{fig:cross_dataset}, both approaches achieve comparable performance on ROC-AUC, PR-AUC, and balanced accuracy on CCLE. These quantitative results are supported by a PCA-based analysis of cellular molecular profiles (Figure~\ref{fig:PCA}), which shows that CTRP--GDSC and CCLE occupy closely overlapping regions in feature space, indicating a relatively limited shift between the two datasets.

Taken together, these findings suggest that when the source and target domains share substantial biological and statistical overlap, unsupervised representation pretraining provides little advantage over standard supervised training. In such settings, tightly coupled representation learning and task supervision appear sufficient to capture the relevant predictive structure. This observation further supports the central premise of this study: the benefits of the proposed staged framework are unlikely to be revealed by cross-dataset cell-line benchmarks alone, but instead emerge under stronger distribution shifts, as examined in the patient-level adaptation experiments that follow.

\begin{figure}[!t]
    \centering
    \includegraphics[width=0.95\linewidth]{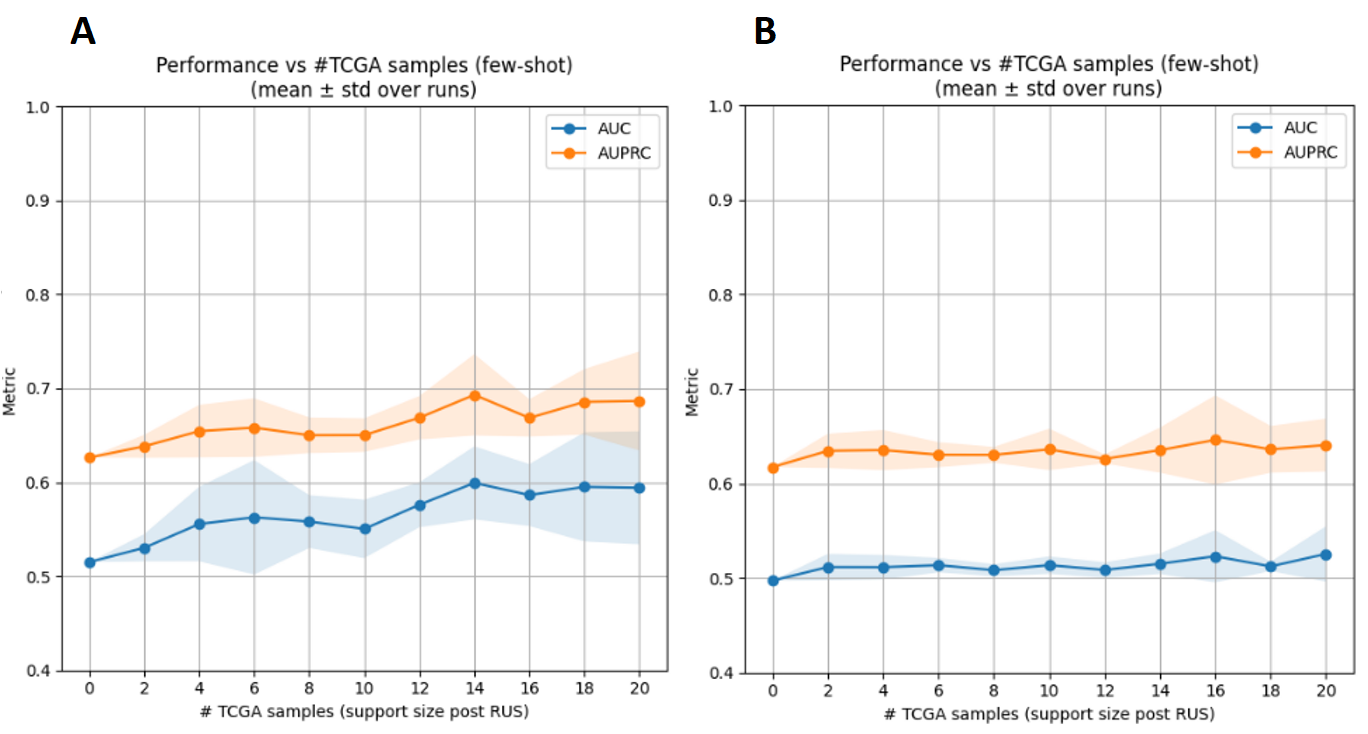}
    \caption{\textbf{Few-shot adaptation to TCGA under strong domain shift.}
ROC--AUC (blue) and area under the precision--recall curve (AUPRC, orange) as a function of the number of labeled TCGA patient samples used for adaptation. (A) STaR-DR. (B) Single-phase baseline (AE-MLP). Curves report the mean over 5 independent runs; shaded bands indicate $\pm$1 standard deviation.}
    \label{fig:Few-shot}
\end{figure}

\subsection{Patient-level adaptation under strong domain shift}

\begin{figure}[t]
    \centering
    \includegraphics[width=0.6\linewidth]{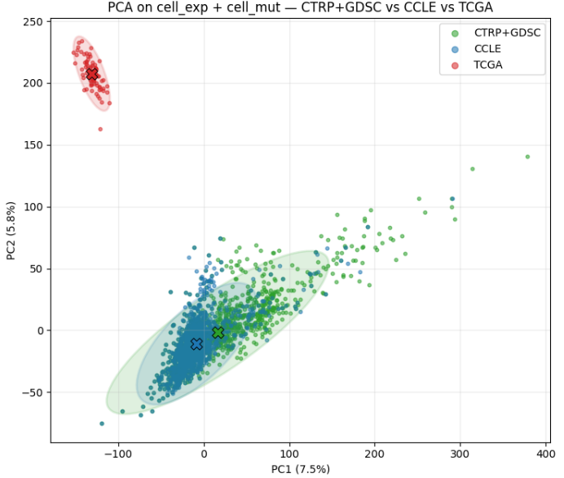}
    \caption{\textbf{PCA of cell-line molecular profiles across datasets.}
    Centroids are connected by Mahalanobis distances computed in PCA space (2D, pooled covariance). Distances are 0.597 between CTRP+GDSC and CCLE, 18.821 between CTRP+GDSC and TCGA, and 18.609 between CCLE and TCGA, based on gene expression and mutation profiles.}
    \label{fig:PCA}
\end{figure}

We finally turn to patient-level transfer on TCGA, which constitutes a substantially more challenging and clinically relevant evaluation setting. As illustrated in Figure~\ref{fig:PCA}, TCGA lies far outside the cell-line manifold, reflecting pronounced biological differences between primary tumors and in vitro cancer models in terms of cellular heterogeneity, microenvironmental context, and clinical confounders. Consistent with this strong domain shift, zero-shot transfer from CTRP--GDSC to TCGA yields weak performance for both methods.

We therefore focus on few-shot adaptation, in which a limited number of labeled TCGA patient--drug pairs are used to fine-tune pretrained models. As shown in Figure~\ref{fig:Few-shot}, predictive performance improves progressively with increasing target-domain supervision for both approaches. However, across all few-shot regimes, the staged framework consistently achieves higher ROC--AUC and PR--AUC than the single-phase baseline.

Strikingly, with as few as 20 labeled TCGA samples, the staged framework already attains substantially higher performance, indicating a faster rate of improvement as a function of target-domain supervision. Importantly, this advantage cannot be attributed to increased model capacity or architectural complexity, as both methods share identical downstream predictors. Instead, the observed gains reflect the quality and transferability of the representations learned during unsupervised pretraining, which enable more efficient specialization to patient-level data under strong biological domain shift.

\subsection{Latent-space analysis and representation quality}

\begin{figure}[t]
    \centering
    \includegraphics[width=0.95\linewidth]{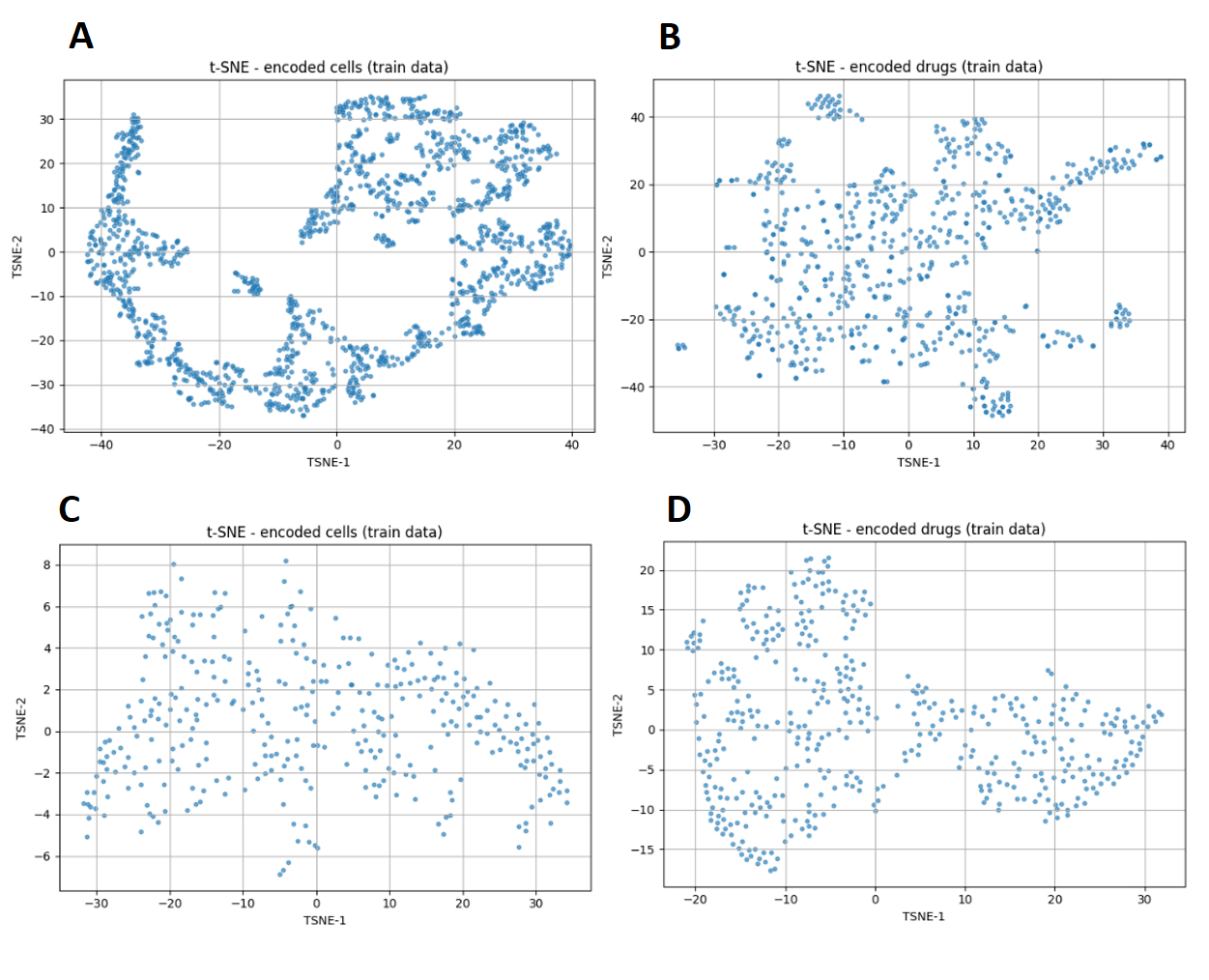}
\caption{\textbf{Latent-space structure of cellular and drug representations.}
t-SNE visualizations of cell and drug embeddings learned by STAR-DR pretraining phase (A)(B) and the single-phase baseline (AE-MLP) (C)(D). Quantitative analysis is performed using the mean $k$-nearest-neighbor radius ($k=10$) and the coefficient of variation, revealing more compact and better organized cellular representations after unsupervised pretraining, while differences in drug embeddings remain limited.}

    \label{fig:representation}
\end{figure}

To gain insight into the observed performance differences, we examine the structure of the learned latent spaces. As shown in Figure~\ref{fig:representation}, cellular embeddings produced by STaR-DR exhibit a more compact and organized structure than those learned by the AE-MLP. This increased compactness and regularity reflect the broader coverage of biological variability acquired during unsupervised pretraining on large collections of unlabeled molecular profiles.

In contrast, differences between drug embeddings remain comparatively modest. This observation is consistent with the more limited diversity of available compound data and with the similar performance of both approaches under the LDO evaluation protocol. Taken together, these results characterize the behavior of the proposed framework across increasing levels of distribution shift.

\section{Discussion}

This study examined the role of unsupervised representation learning in drug-response prediction across progressively stronger domain shifts, with a specific emphasis on sample-efficient adaptation to patient tumors. Rather than seeking improvements in absolute in vitro accuracy, we adopted adaptation efficiency as the primary evaluation criterion, reflecting the constraints of clinical settings where labeled patient data are inherently limited.

\paragraph{Limits of in-domain and cross-dataset benchmarks.}
Across both in-domain evaluations on CTRP--GDSC and cross-dataset generalization to CCLE, STaR-DR did not exhibit systematic advantages over AE-MLP. These findings indicate that when source and target domains share substantial biological and statistical overlap, tightly coupled representation learning and task supervision are sufficient to capture the relevant predictive structure. In such regimes, unsupervised pretraining offers limited additional benefit, suggesting that improvements observed in cell-line benchmarks alone should not be interpreted as evidence of improved clinical transferability.

\paragraph{Adaptation under strong biological domain shift.}
Patient-level transfer to TCGA represents a fundamentally different setting characterized by pronounced biological domain shift. Primary tumors differ from in vitro models along multiple dimensions, including cellular heterogeneity, microenvironmental influences, and clinical confounders, which jointly limit zero-shot generalization from preclinical data. In this regime, STaR-DR exhibits more efficient adaptation when limited labeled patient data become available, highlighting the importance of explicitly separating representation learning from task-specific supervision under clinically realistic conditions.

\paragraph{Mechanistic interpretation.}
Analysis of the learned latent spaces provides insight into why such differences emerge under strong domain shift. As illustrated in Figure~\ref{fig:representation}, unsupervised pretraining yields more compact and structured cellular representations, indicating broader and more regular coverage of biological variability in molecular feature space. These properties facilitate rapid specialization during few-shot adaptation, offering a mechanistic explanation for the improved adaptation efficiency observed at the patient level. In contrast, drug representations exhibit comparatively minor differences between training strategies, consistent with the limited diversity of available compound data and the similar behavior observed under LDO evaluation.

\paragraph{Implications for model evaluation.}
Together, these findings suggest that the practical value of representation learning in DRP cannot be reliably assessed through in-domain or cross-dataset cell-line benchmarks alone. Evaluation strategies focused exclusively on absolute in vitro performance risk underestimating the utility of models designed for clinical translation. By contrast, assessing performance through adaptation efficiency under strong biological domain shift provides a more informative and clinically meaningful perspective, directly aligned with real-world deployment constraints. Accordingly, our comparison against a carefully matched single-phase baseline is sufficient to assess the impact of staged representation learning on adaptation efficiency, independent of architectural complexity or task-specific design choices, and directly aligned with the clinically motivated objective of data-efficient patient-level adaptation.

\paragraph{Limitations and outlook.}
Several limitations remain. The biological gap between cell lines and patient tumors continues to constrain zero-shot transfer, indicating that representation learning alone is insufficient to fully bridge the preclinical-to-clinical divide. Moreover, gains on the drug side remain modest, underscoring the need for richer chemical representations and more diverse compound datasets. Future work may benefit from integrating additional molecular modalities, incorporating lightweight domain-alignment mechanisms, or combining representation learning with causal or mechanistic modeling approaches to further improve robustness under clinical domain shift.

Taken together, these findings indicate that the primary value of unsupervised representation learning for DRP lies not in improving in vitro benchmark performance, but in enabling data-efficient adaptation to patient tumors under strong biological domain shift, thereby substantially reducing the amount of clinical supervision required for effective preclinical-to-clinical translation.

\section{Conclusion}

This work shows that the primary value of unsupervised representation learning in drug-response prediction lies not in improving direct in vitro accuracy, but in enabling sample-efficient adaptation to new biological domains. When training and evaluation domains overlap substantially, as in cell-line benchmarks and closely related pharmacogenomic screens, single-phase supervised models remain sufficient and unsupervised pretraining provides limited additional benefit. In contrast, under strong biological domain shift between in vitro models and patient tumors, explicitly separating representation learning from task supervision yields more transferable representations and supports effective adaptation with very few labeled patient samples.

By leveraging large collections of unlabeled cellular profiles, STaR-DR learns richer and more structured cellular representations that better capture the biological variability encountered in patient tumors. This improved organization facilitates faster adaptation in few-shot regimes, even when zero-shot transfer and in vitro performance remain comparable to existing approaches. Importantly, these gains are achieved without increasing model complexity, underscoring the central role of data efficiency rather than architectural sophistication.

At the same time, important challenges remain. Improvements on the drug side are modest, reflecting the limited diversity of available compound data, and the biological gap between cell lines and patient tumors continues to constrain zero-shot generalization. These observations indicate that representation learning alone cannot fully bridge the preclinical-to-clinical divide and must be complemented by richer chemical information and additional biological context.

Overall, this study argues that evaluating drug-response models through the lens of adaptation efficiency provides a more clinically meaningful perspective than focusing solely on in-domain accuracy. By shifting emphasis from benchmark performance to data-efficient clinical adaptation, staged representation learning offers a practical step toward more realistic and clinically actionable preclinical-to-clinical translation.

\bibliographystyle{vancouver}
\bibliography{literature}

\end{document}